\documentclass{article}
\PassOptionsToPackage{numbers,sort&compress}{natbib}
\usepackage[preprint]{neurips_2026}
\usepackage{tcolorbox} 
\tcbuselibrary{skins}
\usepackage[utf8]{inputenc}
\usepackage[T1]{fontenc}
\usepackage{hyperref}
\usepackage{url}
\usepackage{booktabs}
\usepackage{amsfonts}
\usepackage{amsmath}
\usepackage{amssymb}
\usepackage{nicefrac}
\usepackage{microtype}
\usepackage{graphicx}
\usepackage{doi}
\usepackage{xcolor}
\usepackage{colortbl}
\usepackage{multirow}
\usepackage{subcaption}
\usepackage{algorithm}
\usepackage{algorithmic}
\usepackage{enumitem}
\tcbuselibrary{skins,breakable,listings}
\usepackage{tikz}

\tcbuselibrary{breakable}

\usepackage[textsize=scriptsize, tickmarkheight=0pt]{todonotes}
\usepackage{lipsum} 

\definecolor{todoink}{HTML}{4A5A6A}
\definecolor{todobg}{HTML}{E9ECEF}

\newcommand{\method}{MiA-Signature}

\definecolor{heatlow}{HTML}{FDCFCF} 
\definecolor{heatmid}{HTML}{FFFFCC} 
\definecolor{heathigh}{HTML}{C6EFCE} 
\definecolor{heatbest}{HTML}{7BC67E} 
\definecolor{heatgray}{HTML}{F0F0F0} 

\title{MiA-Signature: Approximating Global Activation for Long-Context Understanding}
\author{
Yuqing Li$^{1,2}$\thanks{Equal contribution.} \;
Jiangnan Li$^{3}$\footnotemark[1] \;
Mo Yu$^{4}$\footnotemark[1] \;
Zheng Lin$^{1,2}$\thanks{Corresponding author.} \; \\
\textbf{Weiping Wang}$^{1}$ \;
\textbf{Jie Zhou}$^{3}$ \\
$^{1}$Institute of Information Engineering, Chinese Academy of Sciences \\
$^{2}$School of Cyber Security, University of Chinese Academy of Sciences \\
$^{3}$Pattern Recognition Center, WeChat AI, Tencent \\
$^{4}$Hunyuan Team, Tencent \\
\texttt{liyuqing@iie.ac.cn}\quad \texttt{\{jiangnanli,moyumyu\}@tencent.com}
}

\hypersetup{
 pdftitle={MiA-Agent},
 pdfkeywords={Agent Memory},
}
\begin{document}
\maketitle

\begin{abstract}
A growing body of work in cognitive science suggests that reportable conscious access is associated with \emph{global ignition} over distributed memory systems, while such activation is only partially accessible as individuals cannot directly access or enumerate all activated contents. This tension suggests a plausible mechanism that cognition may rely on a compact representation that approximates the global influence of activation on downstream processing.
Inspired by this idea, we introduce the concept of \textbf{Mindscape Activation Signature (MiA-Signature)}, a compressed representation of the global activation pattern induced by a query. In LLM systems, this is instantiated via submodular-based selection of high-level concepts that cover the activated context space, optionally refined through lightweight iterative updates using working memory. The resulting MiA-Signature serves as a conditioning signal that approximates the effect of the full activation state while remaining computationally tractable. Integrating MiA-Signatures into both RAG and agentic systems yields consistent performance gains across multiple long-context understanding tasks.



\end{abstract}

\section{Introduction}
\label{sec:intro}

Recent advances in large language models (LLMs) and retrieval-augmented systems have significantly improved performance on knowledge-intensive tasks by combining parametric knowledge with external memory. A dominant paradigm has emerged in which a query is processed, relevant documents are retrieved, and reasoning is performed over the retrieved context. Despite its empirical success, this paradigm implicitly assumes that reasoning can be grounded in a relatively small set of locally retrieved evidence.

However, this assumption appears at odds with insights from cognitive science. A growing body of work suggests that reportable conscious access is associated with \emph{global ignition}—a transient, large-scale activation over distributed memory systems~\cite{dehaene2001towards,tononi2004information,dehaene2011experimental}. At the same time, such activation is only partially accessible: as human beings, we cannot directly access or enumerate all activated contents. Instead, cognition appears to rely on a compact internal representation that approximates the global influence of activation on downstream processing~\cite{tononi2004information,naccache2018and,mashour2020conscious}.

Motivated by this perspective, we argue that memory access in LLM systems can be more effectively modeled as a two-stage process: global activation followed by representation. Rather than directly mapping queries to a small set of retrieved documents, a query first induces a global activation pattern over a semantic memory space, which is then approximated by a tractable representation used to guide downstream computation.

To operationalize this idea, we introduce the notion of a \emph{mindscape}, a global semantic memory space over which activation can be defined. Building on this, we propose the \textbf{Mindscape Activation Signature (MiA-Signature)}, a compressed representation of the activation pattern induced by a query. In practice, MiA-Signatures are constructed via submodular-based selection of high-level concepts that cover the activated context space, optionally refined through lightweight iterative updates using working memory. This representation serves as a conditioning signal that captures a holistic view of relevance, beyond what is available from local retrieval alone.

This perspective leads to a shift in how memory is integrated into reasoning systems. Instead of treating retrieval as the primary interface to memory, we treat activation as the underlying process and signatures as its usable representation. This allows downstream components—such as retrievers, rerankers, or reasoning modules—to operate under a more globally informed semantic context, improving coherence and robustness in long-context settings.

\textbf{Remark: Supporting overcomplete memory. \ }
In realistic settings, memory management systems may produce a large set of memory items, e.g., generated by sleep-time consolidation~\cite{anthropic_claude_code}, sometimes even exceeding the number of raw input items, with substantial redundancy and overlap. By selecting a minimal supporting set that covers the global activation pattern, MiA-Signatures naturally cooperate with such \emph{overcomplete memory}. This allows downstream computation to operate on a holistic approximation of the activated context without incurring the complexity of excessively long inputs recalled from memory.

We evaluate this approach by integrating MiA-Signatures into both retrieval-augmented generation (RAG) pipelines and agentic systems. Empirical results show consistent performance gains across multiple long-context understanding tasks. 
These improvements suggest that approximating global activation provides a more effective interface to memory than relying solely on local retrieval.

In summary, our contributions are as follows:

\begin{itemize}
    \item We introduce a cognitively inspired perspective that models memory access as global activation over a mindscape followed by compact representation.
    \item We propose the \textbf{Mindscape Activation Signature (MiA-Signature)} as a practical instantiation of this idea in LLM systems, providing a compact query-conditioned global state for retrieval, generation, and agentic memory.
    \item We develop a submodular-based construction method, optionally enhanced with lightweight iterative refinement, and demonstrate that integrating MiA-Signatures into both RAG and agentic systems yields consistent improvements on long-context understanding tasks.
\end{itemize}
We believe this work provides a step toward bridging cognitive insights and practical system design, highlighting the importance of global activation in memory-driven reasoning.

\section{Related Work}
\label{sec:related}

\subsection{Evidence Supporting Signatures}

\textbf{Global workspace and global ignition. \ }
The idea that conscious processing involves a form of global information sharing originates from the Global Workspace Theory (GWT)~\cite{baars1988cognitive,baars1997theater}, which proposes that information becomes consciously accessible when it is broadcast to a set of distributed cognitive modules. This framework was later grounded in neurobiological mechanisms through the Global Neuronal Workspace (GNW) theory~\cite{dehaene2001towards,dehaene2005neural,dehaene2011experimental}, which associates conscious access with a nonlinear \emph{global ignition} process---a sudden, large-scale activation sustained by long-range recurrent connectivity. These works establish the existence of global activation as a key substrate of conscious processing.

\textbf{Limits of access and partial awareness. \ }
While GNW posits global activation, subsequent work highlights that such activation is only partially accessible. Recurrent Processing Theory (RPT)~\cite{lamme2006towards} distinguishes between local recurrent processing and global broadcasting, suggesting that not all activated representations reach reportable awareness. Empirical studies on partial awareness and graded consciousness~\cite{kouider2010rich,mashour2020conscious} further support the view that individuals cannot directly access or enumerate all activated contents, even when global activation occurs. These findings point to a gap between the existence of global activation and the form in which it is available for cognition.

\textbf{Integration and compression of global states. \ }
Complementary to GNW, Integrated Information Theory (IIT)~\cite{tononi2004information,tononi2008consciousness} emphasizes that conscious states are highly integrated and structured, rather than collections of independent elements. From this perspective, global brain states are intrinsically compressed representations of distributed activity. Although IIT differs from GNW in its theoretical foundations, both suggest that cognition operates on representations that reflect global structure rather than raw activation patterns.

\textbf{From global activation to usable representations. \ }
Despite these advances, existing theories do not explicitly specify how globally distributed activation is transformed into representations that can guide downstream computation. In parallel, current LLM-based systems, including retrieval-augmented generation (RAG) pipelines, typically access memory through local retrieval mechanisms, implicitly assuming that relevant information can be captured by a small set of retrieved documents. This stands in contrast to the cognitively motivated view that reasoning is shaped by global context.

\textbf{Our perspective. \ }
In this work, we build on these lines of research by proposing that cognition operates on a compact representation that approximates the influence of global activation. We introduce the \textbf{Mindscape Activation Signature (MiA-Signature)} as a computational instantiation of this idea: a compressed representation of a global activation pattern over a semantic memory space. Rather than modeling memory access as direct retrieval, our framework treats it as a two-stage process—global activation followed by signature-based approximation—providing a bridge between cognitive theories of global processing and practical LLM system design.

\subsection{Related Systems: RAG, Memory, and Long-Context Agents}
\label{sec:related:systems}

\textbf{Retrieval as local evidence access. \ }
A dominant line of work improves memory access by making retrieval more iterative, selective, or reasoning-aware, while still treating retrieval itself as the primary interface to external memory. IRCoT~\cite{trivedi2023interleaving} interleaves reasoning with retrieval, and FLARE~\cite{jiang2023active} triggers retrieval when generation becomes uncertain. Self-RAG~\cite{asai2023self}, Adaptive-RAG~\cite{jeong2024adaptive}, and DeepRAG~\cite{guan2025deeprag} further study when and how retrieval should be invoked. More recent systems such as Search-o1~\cite{li2025search} and Search-R1~\cite{jin2025search}expose search as an explicit reasoning action, allowing large reasoning models to interleave thinking with multi-step retrieval and evidence refinement. Despite these advances, the state propagated across steps remains largely local: the current query, reasoning trace, or retrieved passages. Memory access is therefore still framed primarily as iterative evidence lookup rather than as an approximation of a global activated context.

\textbf{Structured retrieval over long documents. \ }
Another line of work improves long-document retrieval by constructing richer
external structures over the source. RAPTOR~\cite{sarthi2024raptor} organizes
documents into a hierarchy of recursive summaries, enabling retrieval at
multiple levels of abstraction. HippoRAG~\cite{gutierrez2024hipporag} builds a
graph-based memory index inspired by hippocampal retrieval. These methods
highlight the importance of global organization for long-context reasoning, moving beyond retrieval over isolated flat chunks. Our work is complementary: rather than treating such structures only as static retrieval substrates, we use them as a mindscape over which a query can induce a compact, query-conditioned activation signature. This signature can then guide retrieval, condition generation, and evolve during multi-step reasoning.

\textbf{Memory-augmented long-context agents. \ }
Recent long-context agents go further by equipping the model with explicit memory states while reading or navigating large inputs. ReadAgent~\cite{lee2024human} compresses long documents into gist memories, and ComoRAG~\cite{wang2026comorag} emphasizes stateful reasoning through a dynamic memory workspace. Moreover,  MemAgent~\cite{yu2025memagent} and ReMemR1~\cite{shi2025look}study how memory can be updated, revisited, or controlled across long reasoning trajectories. These systems are highly relevant to our setting because they move beyond one-shot retrieval to persistent external state. However, their focus is mainly on how to store, revisit, or manage memory during reasoning. Our focus is orthogonal: before local evidence is selected or revisited, we ask how the \emph{global influence} of a query over a semantic memory space can be approximated in a tractable representation.

\section{Method}
\label{sec:method}

We first formalize the mindscape, the query-induced activation pattern, and the MiA-Signature as a compact surrogate of that activation (Sec.~\ref{sec:method:prelim}). We then instantiate the same signature interface in two settings: a static one used once in standard RAG, and a dynamic one maintained as an evolving memory state in an agent loop (Sec.~\ref{sec:method:agent}).

\begin{figure}
    \centering
    \includegraphics[width=1.0\linewidth]{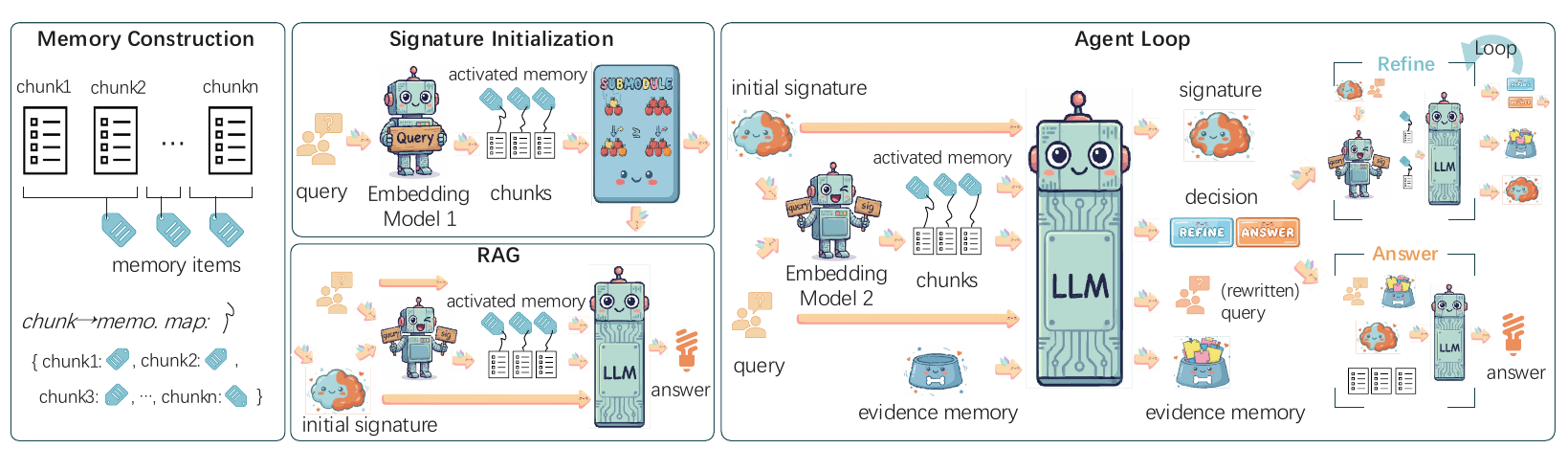}
    \caption{Overview of MiA-Signature. A query first induces a broad activation pattern over the mindscape; MiA-Signature compresses this activated region into a compact, query-conditioned global signal, which then guides retrieval and reasoning in both static RAG and an iterative agent.}
    \label{fig:main}
\end{figure}

\subsection{Preliminaries: MiA-Signature as an Activation Surrogate}
\label{sec:method:prelim}

\begin{tcolorbox}[
  colback=green!6,
  colframe=gray!95,
  boxrule=0.7pt,
  arc=2pt,
  left=6pt,right=6pt,top=5pt,bottom=5pt
]
\textit{A \textbf{MiA-Signature} is a compact, query-conditioned global state that approximates the memory region activated by a query, and that exposes this global signal to downstream retrieval and reasoning.}
\end{tcolorbox}

\subsubsection{Mindscape, Activation, and Signature}

\paragraph{Mindscape.}
Let $D$ denote a long source, such as a novel, a dialogue history, or a document collection. We assume $D$ is associated with a memory pool:
\[
\mathcal{M}(D)=\{m_1,\ldots,m_N\},
\]
where each $m_i$ is grounded in a subset of finer-grained evidence from the source (e.g., passages, chunks).
We refer to this organized memory substrate as the \emph{mindscape}. Memory pools of this kind often contain redundancy, overlap, and multiple levels of abstraction; summaries, extracted entities, and offline-consolidated memories~\cite{anthropic_claude_code} may coexist. This motivates a compact representation of the globally relevant region rather than direct reliance on the full pool.

\paragraph{Activation.}
Given a query $q$, memory access need not be limited to a few locally matched passages. The query typically brings into play a broader semantic region of the mindscape. We represent this query-induced activation as
\begin{equation}
a_q : \mathcal{M}(D) \to \mathbb{R}_{\ge 0},
\label{eq:activation}
\end{equation}
where $a_q(m)$ measures how strongly $m$ belongs to the activated region. In practice, $a_q$ is only approximately observed through retrieval. This is consistent with the broader view that globally activated context may be only partially accessible to downstream processing~\cite{dehaene2011experimental,mashour2020conscious}, and it motivates constructing a compact, usable surrogate of this global signal.

\paragraph{MiA-Signature.}
To make the activation usable, we operate at a higher level of abstraction within the mindscape. Let
$
\mathcal{H}(D)=\{h_1,\ldots,h_M\} \subseteq \mathcal{M}(D)
$
denote a set of high-level memory units---e.g., session summaries or concept-level abstractions---obtained as a coarser-grained projection of $\mathcal{M}(D)$. For a query $q$, let $\mathcal{H}_q \subseteq \mathcal{H}(D)$ be the subset supported by the activated region. We define the MiA-Signature as a compact subset
\begin{equation}
 \sigma^\star(q)
 =
 \arg\max_{\sigma \subseteq \mathcal{H}_q,\, |\sigma|\le K}
 \mathcal{F}\bigl(\sigma;\, q,\, \mathcal{H}_q\bigr),
 \label{eq:signature}
\end{equation}
where $\mathcal{F}$ scores how well a candidate signature serves as a surrogate of the currently activated context, favoring signatures that are relevant to $q$, cover the activated region, and avoid redundancy.

Importantly, $\sigma^\star(q)$ is not intended as a shortened summary of $D$. It is a compact global state that approximates which part of the mindscape has been activated by the query, and it is meant to coexist with locally retrieved evidence rather than replace it. In the agent setting, the signature is further refined as new evidence is consolidated, yielding an evolving global state rather than a one-shot summary~\cite{tononi2004information,naccache2018and}.

\subsubsection{Mindscape-aware Retrieval Interface}
\label{sec:method:prelim-emb}

We use two retrievers with distinct roles, both taken from MiA-RAG~\cite{miarag}. The first, $\mathcal{E}_1$, is a query-only retriever instantiated by SFT-Emb-8B,\footnote{\url{https://huggingface.co/MindscapeRAG/SFT-Emb-8B}} used to obtain an initial view of the relevant memory region before any signature is available. The second, $\mathcal{E}_2$, is a mindscape-aware retriever instantiated by MiA-Emb-8B,\footnote{\url{https://huggingface.co/MindscapeRAG/MiA-Emb-8B}} whose query representation is conditioned on both the input query and a global memory signal. The retriever mechanism is stated in Appendix~\ref{app:retriever_mech}.

In our framework, that global signal is instantiated by the current MiA-Signature $\sigma_t$, so $\mathcal{E}_2$ retrieves with the pair $(q_t,\sigma_t)$: $q_t$ carries the immediate search intent, while $\sigma_t$ supplies the current global memory signal. As $\sigma_t$ evolves, the retrieval distribution evolves with it, letting the system track a changing view of the activated memory region.

\subsection{Instantiating MiA-Signatures in RAG and Agentic Systems}
\label{sec:method:agent}

MiA-Signature provides a common memory interface for two settings. In RAG, the signature is constructed once and used as a fixed conditioning signal. In the agent setting, it is maintained as an evolving global state and updated alongside a local evidence memory as new retrieval steps unfold.

\subsubsection{Step-$0$ Initialization: Submodular Selection for Global Coverage}
\label{sec:method:step0}
Given a query $q$, we first perform a broad retrieval over fine-grained evidence units using the query-only retriever $\mathcal{E}_{1}$. In all experiments, we retrieve the top-$K_0$ candidates with $K_0=50$. Each candidate is then mapped to its associated high-level memory unit, yielding a summary pool:
$
\mathcal{H}_0(q) \subseteq \mathcal{H}(D).
$
This pool provides a coarse, memory-level view of the mindscape region activated by the query, but can be redundant because many retrieved chunks may correspond to overlapping sessions or concepts.

A simple way to construct the initial signature is \emph{First-$K$ truncation}: deduplicate the summaries according to the ranking induced by the step-$0$ retrieval and keep the first $K_{\mathrm{sum}}$. This preserves the local ordering of the initial retriever, but can underrepresent parts of the activated region that appear later in the ranking. We instead select the initial signature with a coverage-aware objective:
\begin{equation}
\sigma_0(q)
=
\arg\max_{\sigma \subseteq \mathcal{H}_0(q),\, |\sigma|\le K_{\mathrm{sum}}}
\mathcal{F}\bigl(\sigma;\, q,\, \mathcal{H}_0(q)\bigr),
\label{eq:init-signature}
\end{equation}
where $\mathcal{F}$ balances query relevance, coverage of the activated region, and diversity among selected memory units. We optimize this set-selection objective with a greedy approximation. Thus, the initial signature is chosen from the same pool as First-$K$, but by how well the selected summaries jointly
represent the activated region rather than by their inherited chunk order. Appendix~\ref{app:submodular} provides the objective, greedy procedure, and comparison with First-$K$ initialization. The resulting $\sigma_0$ serves as the initial MiA-Signature.
\subsubsection{Static Integration: Signature-Augmented RAG}
\label{sec:method:rag}

In the RAG setting, the signature is constructed once and used as a fixed global conditioning signal. Starting from $\sigma_0$, we perform a second retrieval pass with the mindscape-aware retriever $\mathcal{E}_2$. Each candidate evidence unit $c$ is scored by
\begin{equation}
s(c \mid q,\sigma)
=
(1-\alpha)\, s_{\mathrm{qry}}(c \mid q)
+
\alpha\, s_{\mathrm{sig}}(c \mid \sigma),
\label{eq:mia-score}
\end{equation}
where $s_{\mathrm{qry}}(c\mid q)$ measures query relevance,
$s_{\mathrm{sig}}(c\mid\sigma)$ measures consistency with the signature, and
$\alpha\in[0,1]$ controls the strength of the global signal (illustrated in Appendix~\ref{app:retriever_mech}).

The top-$K$ evidence units under this score are passed to the generator. The signature does not replace retrieved evidence; it changes the retrieval interface from query-only matching to query--signature conditioning. When the generator can use global conditioning, $\sigma_0$ is also included in the generation input, either for an LLM with strong context-integration ability or for a smaller mindscape-aware generator trained for this interface, such as MiA-Gen-14B~\cite{miarag}. Thus, static MiA-RAG preserves the efficiency of a two-stage RAG pipeline while exposing a compact approximation of the activated memory region to retrieval, and optionally to generation.
\subsubsection{Dynamic Evolution: Iterative Signature Refinement}
\label{sec:method:agent-loop}

In the agent setting, the same query--signature retrieval interface is reused inside an iterative reasoning loop. Starting from the initial signature $\sigma_0$ in Eq.~\eqref{eq:init-signature}, we set $q_0=q$ and $E_0=\varnothing$. At step $t$, the agent retrieves chunks with the mindscape-aware retriever $\mathcal{E}_2$ conditioned on the current pair $(q_t,\sigma_t)$, using the score in Eq.~\eqref{eq:mia-score}. Let $P_t$ be the retrieved chunks and let $\mathcal{H}_t\subseteq\mathcal{H}(D)$ be the associated high-level memory units.

The state-update model then updates the agent state:
\begin{equation}
(d_t,\;q_{t+1},\;\sigma_{t+1},\;E_{t+1})
=
M_{\mathrm{upd}}(q_t,\sigma_t,P_t,E_t,\mathcal{H}_t),
\label{eq:agent-update}
\end{equation}
where $d_t$ decides whether to answer or continue retrieval. The rewritten query $q_{t+1}$ captures the next local information need, the evidence memory $E_{t+1}$ stores grounded facts accumulated so far, and the refined signature $\sigma_{t+1}$ carries the updated global memory state. The agent therefore
does not rely on query rewriting alone; it navigates long-context memory through the joint evolution of the query, local evidence memory, and global signature.

\subsubsection{Signature-Grounded Answer Generation}
\label{sec:method:generator}

When the agent decides to answer at step $t$, or when the refinement budget is exhausted, the generator receives the original query, the latest retrieved evidence, and the updated memory state:
\begin{equation}
\hat{y}
=
M_{\mathrm{gen}}(q,\;P_t,\;\sigma_{t+1},\;E_{t+1}).
\label{eq:agent-generate}
\end{equation}
Generation remains grounded in local evidence while using the refined signature as the compact global state produced by the loop.

\begin{algorithm}[t]
\small
\caption{\method{} agent over a long source}
\label{alg:agent}
\begin{algorithmic}[1]
\REQUIRE Query $q$; source $D$ with memory pool $\mathcal{M}(D)$ and high-level memory set $\mathcal{H}(D)$; stopping budget $N_{\mathrm{stop}}$; query-only retriever $\mathcal{E}_1$; mindscape-aware retriever $\mathcal{E}_2$; update model $M_{\mathrm{upd}}$; generator $M_{\mathrm{gen}}$.
\STATE $\sigma_0 \gets \textsc{InitSignature}(q, D; \mathcal{E}_1)$ \hfill // Eq.~\eqref{eq:init-signature}
\STATE $q_0 \gets q$; \quad $E_0 \gets \varnothing$
\FOR{$t = 0$ \TO $N_{\mathrm{stop}}-1$}
    \STATE $P_t \gets \textsc{Retrieve}(q_t, \sigma_t; \mathcal{E}_2)$ \hfill // Eq.~\eqref{eq:mia-score}
    \STATE $\mathcal{H}_t \gets \textsc{Summaries}(P_t)$
    \STATE $(d_t, q_{t+1}, \sigma_{t+1}, E_{t+1}) \gets \textsc{Update}(q_t, \sigma_t, P_t, E_t, \mathcal{H}_t; M_{\mathrm{upd}})$
    \IF{$d_t = \textsc{Answer}$}
        \RETURN $M_{\mathrm{gen}}(q, P_t, \sigma_{t+1}, E_{t+1})$
    \ENDIF
\ENDFOR
\RETURN $M_{\mathrm{gen}}(q, P_{N_{\mathrm{stop}}-1}, \sigma_{N_{\mathrm{stop}}}, E_{N_{\mathrm{stop}}})$
\end{algorithmic}
\end{algorithm}

\section{Experiments}
\subsection{Experimental Setup}
\label{sec:setup}
We evaluate MiA-Signatures in two long-context memory-access settings:
a static RAG pipeline and an iterative agent. The static setting tests a
one-shot signature as a compact global conditioning signal, while the agent
setting tests whether the same interface remains useful as an evolving memory state over multiple retrieval steps.

\subsubsection{Datasets and Metrics}
\label{sec:setup:data-metrics}

We evaluate on four long-context benchmarks covering multiple-choice QA, open-ended QA, multi-hop QA, and claim verification. DetectiveQA~\cite{xu2025detectiveqa}
evaluates multiple-choice reasoning over detective novels in English and Chinese. NarrativeQA~\cite{kovcisky2018narrativeqa} evaluates open-ended question answering over narrative texts. NovelHopQA~\cite{gupta2025novelhopqa} evaluates multi-hop reasoning over long novel excerpts, and NoCha~\cite{nocha} evaluates claim verification over full novels.

For DetectiveQA and NarrativeQA, we adopt a \emph{series-book construction}. Instead of treating each novel as an independent source, we merge books from the same series into a single long document, e.g., Agatha Christie's \emph{Miss Marple} and \emph{Hercule Poirot} series for DetectiveQA. 
The questions remain tied to episode-specific evidence, but retrieval is performed over a larger memory
space containing related characters, events, and distractors.
Appendix~\ref{app:datasets:series} details the aggregation procedure, and Appendix~\ref{app:datasets:single-vs-series} provides a single-book vs.\ series-book comparison showing such retrieval interference.

We use accuracy for multiple-choice QA, F1 score for open-ended QA, and accuracy together with pair accuracy for NoCha. We also report Recall@10 when gold evidence annotations are available.

\subsection{Implementation Details}
\label{sec:setup:impl}
Unless otherwise specified, the agent uses DeepSeek-V3.2~\cite{liu2025deepseek} as both the state update model $M_{\mathrm{upd}}$ and the final answer generator $M_{\mathrm{gen}}$. The agent runs for at most three refinement steps. At step $0$, the query-only retriever returns 50 candidate chunks; these chunks are mapped to high-level memory units, from which at most five session summaries are selected to form the initial signature. Each subsequent retrieval step returns 20 chunks. The dual-signal retrieval score uses $\alpha=0.5$ to balance query relevance and signature consistency.

The high-level memory set $\mathcal{H}(D)$ is constructed offline by splitting each document into source-order windows of $W=20$ chunks and summarizing each window once using GPT-4o with a fixed summary-construction prompt. The resulting session summaries are cached, query-independent, and reused across all queries over the same document. Appendix~\ref{app:submodular:setup} gives the formal chunk-to-summary mapping.
Submodular selection follows Appendix~\ref{app:submodular}. In the static RAG setting, the coverage-aware variant scores all three terms with BGE-M3~\cite{bge-m3} \textsc{CLS} embeddings, using default weights $(\lambda_Q,\lambda_C,\lambda_D)=(0.3,0.4,0.3)$. The First-$K$ variant used by the agent reuses step-$0$ retriever scores directly and invokes no additional encoder.

\subsubsection{Baselines}
\label{sec:setup:methods}

We compare two families of systems: static RAG pipelines and iterative agents. The RAG experiments isolate where the MiA-Signature is used in a retriever--generator pipeline, while the agent experiments test whether the signature remains useful as an evolving memory state.

\textbf{RAG methods. \ }
Each RAG system is reported as a retriever--generator pair.
\textbf{Query-only RAG} retrieves with the input query alone and does not use a
MiA-Signature. We evaluate three query-only variants: \textbf{Qwen3-Emb /
Qwen-14B} as a 14B-scale reference, \textbf{Qwen3-Emb / DS-V3.2} as a stronger generator baseline, and \textbf{MiA-Emb / DS-V3.2} as a retriever-backbone control without signature conditioning.
In the \textbf{MiA-Emb} configuration, the MiA-Signature is used to condition
retrieval, while the generator receives only the retrieved chunks.
\textbf{MiA-RAG} further provides the same signature to the generator, forming
the full signature-aware RAG interface. We evaluate MiA-RAG with both DS-V3.2
and MiA-Gen-14B~\cite{miarag}. All static signature-based methods use the same coverage-aware submodular step-$0$ initialization (\S~\ref{sec:method:step0}).

\textbf{Agent methods. \ }
All agent variants start from the same broad step-$0$ retrieval and use at most three refinement steps. \textbf{Agent w/o Sig.} follows the same iterative retrieval process as MiA-Agent but removes the signature from the agent state. We report two answer-time inputs for this baseline: retrieved chunks only, and retrieved chunks plus accumulated evidence memory (\textsc{Evi.}).
\textbf{MiA-Agent} maintains an evolving signature $\sigma_t$ and retrieves with $(q_t,\sigma_t)$ at each step. To separate retrieval-time and generation-time effects, we vary the final generator input. All variants use the final retrieved chunks; we additionally provide the final signature (\textsc{Sig.}), accumulated evidence memory (\textsc{Evi.}), or both (\textsc{Sig.+Evi.}). These variants test whether the evolving signature only steers retrieval or also helps answer generation. MiA-Agent initializes $\sigma_0$ with the lightweight First-$K$ submodular variant, as it is later refined online.

\begin{table*}[t]
\centering
\scriptsize
\setlength{\tabcolsep}{3pt}
\caption{
\textbf{RAG results.}
MiA-Emb uses the MiA-Signature only for retrieval: the retriever is conditioned on both the query and the signature, while the generator receives retrieved chunks only. MiA-RAG uses the full signature-aware interface, where the same signature is used by both the retriever and the generator.  Avg.\ Perf.\ averages the main task metric of each benchmark, using PairAcc for NoCha. Best final-task results are in \textbf{bold}.
}
\label{tab:comprehensive-rag-results}

\resizebox{\textwidth}{!}{
\begin{tabular}{lll|cc|cc|cc|c|cc}
\toprule
\multirow{2}{*}{Method}
& \multirow{2}{*}{Retriever}
& \multirow{2}{*}{Generator}
& \multicolumn{2}{c|}{DetectiveQA (EN/ZH)}
& \multicolumn{2}{c|}{NarrativeQA}
& \multicolumn{2}{c|}{NovelHopQA}
& NoCha
& \multicolumn{2}{c}{Avg.} \\
& & 
& $R@10$ & Acc
& $R@10$ & F1
& $R@10$ & F1
& Acc / PairAcc
& $R@10$ & Perf. \\
\midrule
Query-only RAG
& Qwen3-Emb
& Qwen-14B
& 24.7 / 29.3 & 50.7 / 56.7
& 48.5 & 36.6
& 34.3 & 35.8
& 64.3 / 31.8
& 36.6 & 39.5 \\

Query-only RAG
& Qwen3-Emb
& DS-V3.2
& 24.7 / 29.3 & 58.7 / 68.0
& 48.5 & 41.8
& 33.7 & 37.0
& 74.6 / 49.2
& 36.4 & 47.8 \\

Query-only RAG
& MiA-Emb
& DS-V3.2
& 34.0 / 42.7 & 59.3 / 76.0
& 51.3 & 41.1
& 36.6 & 38.0
& 81.0 / 61.9
& 42.1 & 52.2 \\
\midrule
\rowcolor{blue!8}
MiA-Emb
& MiA-Emb (+sig)
& DS-V3.2
& 44.7 / 42.7 & 70.7 / 78.0
& 59.5 & 45.1
& 36.8 & 38.5
& 79.4 / 58.7
& 46.7 & 54.2 \\

\rowcolor{blue!8}
MiA-RAG
& MiA-Emb (+sig)
& MiA-Gen-14B (+sig)
& 44.7 / 42.7 & 70.7 / 73.3
& 59.5 & \textbf{48.0}
& 36.8 & 27.9
& 71.4 / 47.6
& 46.7 & 48.9 \\

\rowcolor{blue!8}
\textbf{MiA-RAG}
& \textbf{MiA-Emb (+sig)}
& \textbf{DS-V3.2 (+sig)}
& 44.7 / 42.7 & \textbf{74.7 / 80.0}
& 59.5 & 42.8
& 36.8 & \textbf{38.7}
& \textbf{82.5 / 65.1}
& 46.7 & \textbf{56.0} \\
\bottomrule
\end{tabular}
}
\end{table*}

\subsection{Main Results}
\label{sec:results}

We organize the experiments around three questions. 

\paragraph{RQ1: Does conditioning retrieval on a MiA-Signature improve static RAG?}
Table~\ref{tab:comprehensive-rag-results} first evaluates whether the
MiA-Signature helps static RAG at the retrieval stage. The Qwen3-Emb rows serve as general query-only baselines: they retrieve with the input query alone and do not have a mechanism for using a global memory state. Previous studies also show that simply appending a summary to a general embedding model can hurt retrieval, as the added global context may blur the query focus rather than guide selection~\cite{miarag}. This motivates a signature-aware retriever rather than a query-plus-summary shortcut.

Under the same retriever and generator backbone, conditioning retrieval on the MiA-Signature improves average $R@10$ by 10.9\% and average task performance by 3.8\%. Since the generator input remains the retrieved chunks only, the gain comes from changing how evidence is selected before generation, rather than from giving the generator more context.
The improvement is most meaningful on DetectiveQA and NarrativeQA, where the answer often depends on a dispersed region of related events, entities, or claims. In such cases, query-only retrieval can find locally plausible chunks while missing the broader semantic region; the signature helps reduce this mismatch. NovelHopQA shows a smaller gain, marking a boundary of this mechanism: the signature helps locate a relevant semantic region, but multi-hop questions still require composing specific evidence chains that a compact global state may not fully specify.
These results support MiA-Signature as a retrieval-side memory interface. It does not replace local evidence; it changes how local evidence is selected.

\paragraph{RQ2: Does the signature remain useful as memory access becomes iterative?}
Table~\ref{tab:main-agent} evaluates whether the retrieval-side benefit of the MiA-Signature extends from static RAG to an iterative agent. Compared with \textbf{Agent w/o Sig.}, \textbf{MiA-Agent} improves retrieval recall on every benchmark with retrieval annotations, with the clearest gains on DetectiveQA-ZH and NovelHopQA. Compared with the static MiA-RAG reference, MiA-Agent largely matches or improves retrieval despite starting from a lightweight First-$K$ signature, suggesting that iterative signature updates can compensate for a simpler initial state.
This matters because iterative retrieval can otherwise become overly tied to the current rewritten query. As the agent accumulates evidence, the local query may narrow or drift, while the original problem may still require a broader activated memory region. MiA-Agent addresses this by maintaining an evolving signature $\sigma_t$ alongside the query and a working evidence memory. The signature guides search at the global level, while the evidence memory preserves grounded facts already retrieved.

These results extend the retrieval-side conclusion from RQ1: the MiA-Signature is not only useful as a one-shot retrieval-conditioning signal, but also as a stable global state that keeps iterative search aligned with the query-induced activated region across steps. The two memory states are not interchangeable; we analyze their answer-time effects in RQ3.

\paragraph{RQ3: Which memory state should be exposed to the final generator?}
The answer-time ablations show that retrieval-time and generation-time uses of memory should be separated. In static RAG, \textbf{MiA-RAG} improves over \textbf{MiA-Emb}, indicating that the signature can provide useful global context to the generator in addition to guiding retrieval. However, this benefit is not automatic. The MiA-Gen-14B variant achieves the best NarrativeQA F1, but does not dominate across benchmarks. This suggests that answer-time use of the signature depends on both the task and the generator's ability to exploit it. Moreover, the agent ablations make this distinction clearer. The final signature and the working evidence memory encode different types of information. The signature summarizes the broader activated memory region, while the evidence memory preserves grounded facts accumulated during the agent loop. On NoCha, where local factual continuity is important, exposing both states gives the best result. By contrast, on NarrativeQA and NovelHopQA, the best MiA-Agent variants use retrieved chunks alone. Once the retrieved chunks already contain a usable answer path, additional memory state may distract the generator rather than provide useful structure.

Taken together, retrieval benefits from the signature more consistently than generation does. The signature is a reliable search-guiding state, but its answer-time value is selective. It helps when global constraints are needed to interpret local evidence, and it can be unnecessary when the retrieved chunks already provide a direct and composable evidence path.

\begin{table*}[t]
\centering
\scriptsize
\setlength{\tabcolsep}{3pt}
\caption{
\textbf{Agent results and answer-time ablation.}
All iterative agent variants use DeepSeek-V3.2 with a three-step refinement
budget. The static MiA-RAG row is included as a non-iterative reference using
the same generator. All answer-time inputs include the final retrieved chunks;
\textsc{Sig.} denotes the final MiA-Signature, and \textsc{Evi.} denotes the
accumulated evidence memory.
}
\label{tab:main-agent}

\resizebox{\textwidth}{!}{
\begin{tabular}{ll|cc|cc|cc|cc}
\toprule
\multirow{2}{*}{System} & \multirow{2}{*}{Answer-time input}
& \multicolumn{2}{c|}{DetectiveQA (EN/ZH)}
& \multicolumn{2}{c|}{NarrativeQA}
& \multicolumn{2}{c|}{NovelHopQA}
& \multicolumn{2}{c}{NoCha} \\
& & $R@10$ & Acc & $R@10$ & F1 & $R@10$ & F1 & Acc & PairAcc \\
\midrule

\rowcolor{gray!15}
Agent w/o Sig. & Chunks
& 42.7 / 46.7 & 68.0 / 82.0
& 53.7 & 42.4
& 33.9 & 37.4
& 77.8 & 57.1 \\

\rowcolor{gray!15}
Agent w/o Sig. & Chunks + \textsc{Evi.}
& 42.7 / 46.7 & 76.0 / 80.0
& 53.7 & 43.4
& 33.9 & 36.4
& 84.9 & 69.8 \\

MiA-RAG (static) & Chunks + \textsc{Sig.}
& 44.7 / 42.7 & 74.7 / 80.0
& \text{59.5} & 42.8
& 36.8 & \textbf{38.7}
& 82.5 & 65.1 \\

\midrule

MiA-Agent & Chunks
& \text{46.7 / 52.7} & 68.7 / 81.3
& 59.1 & \textbf{45.3}
& \text{39.3} & \textbf{38.7}
& 80.2 & 61.9 \\

MiA-Agent & Chunks + \textsc{Sig.}
& \text{46.7 / 52.7} & \textbf{76.7} / 82.0
& 59.1 & 44.9
& \text{39.3} & 37.1
& 83.3 & 68.3 \\

MiA-Agent & Chunks + \textsc{Evi.}
& \text{46.7 / 52.7} & 73.3 / \textbf{86.0}
& 59.1 & 43.6
& \text{39.3} & 36.2
& 83.3 & 66.7 \\

MiA-Agent & Chunks + \textsc{Sig.} + \textsc{Evi.}
& \text{46.7 / 52.7} & 73.3 / 80.0
& 59.1 & 44.3
& \text{39.3} & 35.6
& \textbf{85.7} & \textbf{71.4} \\

\bottomrule
\end{tabular}
}
\end{table*}

\vspace{-2mm}

\subsection{Analysis}
\label{sec:analysis}
We include two targeted studies to further examine the mechanism behind the main results. First, Appendix~\ref{app:submod-vs-firstk} compares the two
submodular initializers, Coverage-aware and First-$K$, under the static RAG pipeline. Both variants use the same step-$0$ candidate pool, so the comparison isolates whether coverage-aware selection provides benefit beyond simply taking
the ranking prefix. Second, Appendix~\ref{app:coevo-ablation} studies query rewriting in the agent loop. We find that rewriting is best treated as a control knob rather than the core mechanism. It helps when refinement should narrow the search, but can be harmful when the task requires preserving multiple evidence paths. Accordingly, we keep the query fixed on NovelHopQA and rewrite it on the other benchmarks.
Finally, the case study in Appendix~\ref{app:case-study} illustrates the same division of labor observed in the aggregate results: local chunks provide grounded evidence, working evidence memory preserves accumulated facts across steps, and the MiA-Signature maintains a compact global state that keeps retrieval and generation aligned with the activated memory region.
\vspace{-2mm}
\section{Conclusion}
\label{sec:conclusion}

We introduced MiA-Signature, a compact representation of the global activation pattern induced by a query over a structured memory space. This representation serves as a tractable interface between broad memory activation and downstream LLM computation.
We instantiate this idea in both static RAG and agentic systems, showing that a compact activation signature can improve how LLMs access and use external memory across different inference settings. Across long-context benchmarks, MiA-Signatures consistently improve over query-only counterparts. The results suggest that compact representations of global activation can provide useful memory context without replacing local evidence or requiring direct access to the full activated memory state.
These findings support a view of memory access in LLM systems as global activation followed by compact representation. MiA-Signature offers one practical step toward this interface, connecting distributed memory activation with local evidence-based reasoning.

\bibliographystyle{plainnat}
\bibliography{references}

%
%
%
\appendix
\section{Submodular Initialization: Coverage-aware vs.\ First-$K$}
\label{app:submodular}

This appendix details the submodular selection framework used to construct the initial MiA-Signature at step~$0$. We consider two variants: \textbf{Coverage-aware submodular}, which uses query relevance, chunk-level coverage, and diversity, and \textbf{First-$K$ submodular}, a relevance-only degenerate variant that is modular and therefore trivially submodular. Static RAG uses the Coverage-aware variant, while the agent uses the First-$K$ variant because the signature is later refined online. We formalize the objective, discuss its submodularity, describe the algorithm, and compare the two variants under identical retrieval pipelines.

\subsection{Problem Setup and Motivation}
\label{app:submodular:setup}

Let $q$ denote the query and let $\mathcal{C} = (c_1, c_2, \dots, c_M)$ be the
rank-ordered list of candidate chunks returned by the step-$0$ query-only
retriever (cf.\ Sec.~\ref{sec:method}). Each chunk $c_i$ is associated with a
\emph{session summary} $s_{\pi(i)} \in \mathcal{S} = \{s_1, \dots, s_N\}$ via a
deterministic mapping $\pi : \mathcal{C} \to \mathcal{S}$ induced by document
sessionization. Distinct chunks may share a session summary, so typically
$N \ll M$. Our goal is to select a subset
$\mathcal{A} \subseteq \mathcal{S}$ with $|\mathcal{A}| \leq K$ whose
concatenation forms the initial signature $\sigma_0$ fed to the retriever at
step~$1$.
\paragraph{Session summaries.}
We construct session summaries once for each document $D$ before any query is
issued. Let $\mathcal{X}(D)=(x_1,\ldots,x_L)$ denote the source-order chunk
sequence of $D$. We divide this sequence into non-overlapping contiguous
windows of $W=20$ chunks and summarize each window with a single GPT-4o~\cite{hurst2024gpt} call using a fixed summary-construction prompt~\cite{li2026query}. The full prompt is given in Appendix~\ref{app:prompts}.

The resulting summaries form
\[
    \mathcal{S}=\{s_1,\ldots,s_J\}, \qquad
    J=\left\lceil \frac{L}{W} \right\rceil .
\]
In our experiments, we instantiate the high-level memory set as
$\mathcal{H}(D)\equiv\mathcal{S}$. The chunk-to-summary mapping is fixed by
the source-order window assignment:
\[
    \pi(x_\ell) = s_{\lceil \ell/W\rceil}.
\]
For a retrieved candidate chunk $c$, $\pi(c)$ denotes the cached session
summary assigned to the source chunk from which $c$ originates. This
construction is deterministic, query-independent, and fixed at indexing time.
We use the same window size, GPT-4o summarizer, and summary-construction prompt across all datasets. Since $\mathcal{S}$ is cached, summary construction adds no query-time LLM calls and the summaries are reused across all queries over $D$.

\paragraph{Why not First-$K$?}
The simple variant selects
$\mathcal{A} = \{s_{\pi(i)}\}_{i=1}^{K}$, the session summaries associated with
the top-$K$ chunks. This is efficient but can be redundant, since multiple
top-ranked chunks may map to the same session. It can also miss useful
summaries just below the rank cutoff, and the chunk-level ranking may not
reflect summary-level alignment with the overall information need.

Coverage-aware submodular selection addresses these issues by adding chunk-level coverage and diversity to the relevance-only objective.
\subsection{Objective Formulation}
\label{app:submodular:obj}

Let $\mathbf{e}_q \in \mathbb{R}^d$, $\mathbf{e}_{s_j} \in \mathbb{R}^d$,
$\mathbf{e}_{c_i} \in \mathbb{R}^d$ be $\ell_2$-normalized BGE-M3 \textsc{cls}
embeddings of the query, summary $s_j$, and chunk $c_i$, respectively. Define:

\paragraph{(i) Query relevance.} The sum of cosine similarities between the
query and the selected summaries:
\begin{equation}
    f_Q(\mathcal{A}) \;=\; \sum_{s \in \mathcal{A}} \mathbf{e}_q^{\top} \mathbf{e}_{s}
    \label{eq:fq}
\end{equation}

\paragraph{(ii) Chunk coverage.} For a candidate chunk $c_i$ of rank $r_i$
(with $r_1 = 1$), assign a rank-decaying weight $w_i = 1/(r_i + 1)$.
Define the \emph{match score} between summary $s$ and chunk $c$ as
$m(s, c) = \max(0,\, \mathbf{e}_s^{\top} \mathbf{e}_c)$. Let $\mathrm{Cov}(s) \subseteq
\mathcal{C}$ denote the chunks whose session summary equals $s$
(i.e., $\mathrm{Cov}(s) = \pi^{-1}(s)$). The coverage term is:
\begin{equation}
    f_C(\mathcal{A}) \;=\; \sum_{c_i \in \mathcal{C}} w_i \cdot
        \max_{s \in \mathcal{A}, \, c_i \in \mathrm{Cov}(s)}
        m(s, c_i),
    \label{eq:fc}
\end{equation}
with the convention that the inner $\max$ is zero when no selected summary
covers $c_i$. Two compounding weights are at play: $w_i$ biases toward
chunks highly ranked by the step-$0$ retriever, while $m(s, c_i)$ is a semantic
fidelity check that a chunk is genuinely reflected in its session summary.

\paragraph{(iii) Diversity.} Penalize angular similarity to the existing
selection:
\begin{equation}
    f_D(s \mid \mathcal{A}) \;=\;
    \begin{cases}
        1 & \mathcal{A} = \varnothing \\
        1 - \max_{s' \in \mathcal{A}} \mathbf{e}_s^{\top} \mathbf{e}_{s'}
          & \text{otherwise}
    \end{cases}
    \label{eq:fd}
\end{equation}

\paragraph{Combined marginal gain.} Letting $\tilde{f}_Q, \tilde{f}_C$ denote
max-normalized variants of $f_Q, f_C$ (to place the three terms on a common
scale), the marginal gain of adding summary $s$ to the current selection
$\mathcal{A}$ is:
\begin{equation}
    \Delta(s \mid \mathcal{A}) =
    \lambda_Q \,\Delta \tilde{f}_Q(s \mid \mathcal{A}) +
    \lambda_C \,\Delta \tilde{f}_C(s \mid \mathcal{A}) +
    \lambda_D \, f_D(s \mid \mathcal{A}),
    \label{eq:gain}
\end{equation}
with default weights $\lambda_Q = 0.3$, $\lambda_C = 0.4$, $\lambda_D = 0.3$.

\subsection{Submodularity Analysis}
\label{app:submodular:proof}

Recall that a set function $f: 2^{\mathcal{S}} \to \mathbb{R}$ is
\emph{monotone} if $f(\mathcal{A}) \leq f(\mathcal{B})$ whenever
$\mathcal{A} \subseteq \mathcal{B}$, and \emph{submodular} if for all
$\mathcal{A} \subseteq \mathcal{B} \subseteq \mathcal{S}$ and
$s \notin \mathcal{B}$,
\[
    f(\mathcal{A} \cup \{s\}) - f(\mathcal{A})
    \;\geq\;
    f(\mathcal{B} \cup \{s\}) - f(\mathcal{B})
\]
(the diminishing-returns property).

\paragraph{Proposition 1 ($f_Q$ is modular).} The query-relevance term $f_Q$
is a sum of element-wise quantities $\mathbf{e}_q^{\top} \mathbf{e}_s$ and is
therefore \emph{modular}, hence trivially submodular and monotone. 


\paragraph{Proposition 2 ($f_C$ is monotone submodular).} The coverage term
$f_C$ in Eq.~\ref{eq:fc} has the form of a weighted \textsc{max-coverage}:
\[
    f_C(\mathcal{A}) = \sum_{c_i \in \mathcal{C}} w_i \cdot
        \max_{s \in \mathcal{A}} \big[\, m(s, c_i) \cdot
        \mathbf{1}[c_i \in \mathrm{Cov}(s)] \,\big].
\]
Each summand is a weighted maximum of non-negative quantities over
$\mathcal{A}$; the $\max$ operator over a growing set is monotone and
submodular, and non-negative linear combinations preserve both properties
\cite{Nemhauser1978AnAO}. 


\paragraph{Status of $f_D$.} The diversity term $f_D$ is \emph{not} monotone:
Adding a summary $s$ similar to an existing selection can decrease its value.
Consequently $\Delta(\cdot \mid \cdot)$ in Eq.~\ref{eq:gain} is not globally
submodular. This is a deliberate design choice: the diversity term mainly
acts as a tie-breaker between candidates with similar $f_Q/f_C$ profiles,
and the $(1 - 1/e)$ approximation guarantee of greedy maximization under
\emph{monotone} sub-modular objectives \citep{Nemhauser1978AnAO} still
holds for the $f_Q + f_C$ sub-objective obtained by setting $\lambda_D = 0$.

\paragraph{Approximation guarantee.} Ignoring the diversity term, the
remaining objective $f_{QC}(\mathcal{A}) = \lambda_Q \tilde{f}_Q(\mathcal{A}) +
\lambda_C \tilde{f}_C(\mathcal{A})$ is monotone submodular and non-negative.
The classical result of Nemhauser, Wolsey, and Fisher
\citep{Nemhauser1978AnAO} guarantees that greedy selection returns a
solution $\mathcal{A}_\mathrm{greedy}$ such that
\begin{equation}
    f_{QC}(\mathcal{A}_\mathrm{greedy})
    \;\geq\; \left(1 - \tfrac{1}{e}\right) \cdot
    f_{QC}(\mathcal{A}^{*}),
\end{equation}
where $\mathcal{A}^{*}$ is the optimal size-$K$ subset. This
$(1 - 1/e) \approx 0.632$ bound applies to the dominant monotone component
of our objective.

\subsection{Algorithmic Procedure}
\label{app:submodular:alg}

We implement greedy maximization of Eq.~\ref{eq:gain} (Algorithm~\ref{alg:submod}),
which expands the \textsc{InitSignature} call at line~1 of the main-paper
agent loop (Alg.~\ref{alg:agent} in Sec.~\ref{sec:method:step0}).
Embeddings are computed once per call in a single batched forward pass through
BGE-M3 \citep{bge-m3}, then cached. Complexity is
$\mathcal{O}(K \cdot N \cdot M)$ set operations after the embedding pass,
which is negligible relative to the retriever call itself.

\begin{algorithm}[h]
\caption{Coverage-Aware Sub-modular Summary Selection}
\label{alg:submod}
\begin{algorithmic}[1]
\REQUIRE Query $q$; candidate chunks $\mathcal{C}$ with ranks;
         summary set $\mathcal{S}$; coverage map $\pi^{-1}$;
         budget $K$; weights $\lambda_Q, \lambda_C, \lambda_D$
\ENSURE Selected summaries $\mathcal{A}$, size $\leq K$
\STATE Compute embeddings $\mathbf{e}_q, \{\mathbf{e}_s\}, \{\mathbf{e}_c\}$ (batched BGE-M3, CLS, $\ell_2$-norm)
\STATE $w_i \gets 1 / (r_i + 1)$ for each $c_i \in \mathcal{C}$
\STATE Pre-compute $q_s \gets \mathbf{e}_q^{\top} \mathbf{e}_s$ for all $s \in \mathcal{S}$
\STATE Pre-compute $\mathrm{cov}_s \gets \sum_{c \in \pi^{-1}(s)} w_c \cdot m(s, c)$
\STATE Normalize: $\tilde{q}_s \gets q_s / \max_{s'} q_{s'}$;\quad $\tilde{\mathrm{cov}}_s \gets \mathrm{cov}_s / \max_{s'} \mathrm{cov}_{s'}$
\STATE $\mathcal{A} \gets \varnothing$;\quad $\mathrm{covered}(c) \gets 0$ for all $c \in \mathcal{C}$
\FOR{$k = 1$ \TO $K$}
    \STATE $\Delta^{\star} \gets -\infty$,\quad $s^{\star} \gets \texttt{None}$
    \FORALL{$s \in \mathcal{S} \setminus \mathcal{A}$}
        \STATE $\Delta_Q \gets \tilde{q}_s$ \hfill \COMMENT{modular, Eq.~\ref{eq:fq}}
        \STATE $\Delta_C \gets \tfrac{1}{Z_C}
\sum_{c \in \pi^{-1}(s)} w_c \cdot m(s,c) \cdot (1-\mathrm{covered}(c))$ \hfill \COMMENT{marginal coverage, Eq.~\ref{eq:fc}}
        \STATE $\Delta_D \gets 1 - \max_{s' \in \mathcal{A}} \mathbf{e}_s^{\top} \mathbf{e}_{s'}$, or $1$ if $\mathcal{A} = \varnothing$ \hfill \COMMENT{Eq.~\ref{eq:fd}}
        \STATE $\Delta \gets \lambda_Q \Delta_Q + \lambda_C \Delta_C + \lambda_D \Delta_D$
        \IF{$\Delta > \Delta^{\star}$}
            \STATE $\Delta^{\star} \gets \Delta$,\quad $s^{\star} \gets s$
        \ENDIF
    \ENDFOR
    \STATE \textbf{if} $s^{\star} = \texttt{None}$ \textbf{then} \textbf{break}
    \STATE $\mathcal{A} \gets \mathcal{A} \cup \{s^{\star}\}$
    \STATE For each $c \in \pi^{-1}(s^{\star})$: $\mathrm{covered}(c) \gets 1$
\ENDFOR
\RETURN $\mathcal{A}$
\end{algorithmic}
\end{algorithm}

\paragraph{Implementation note.} The normalization constant $Z_C$ in line~11
is $\max_{s'} \mathrm{cov}_{s'}$ (the same denominator used in line~5),
which keeps the three $\Delta$-terms on a comparable $[0, 1]$ scale. The
BGE-M3 model is loaded once per process and cached, so the algorithm adds
no overhead when submodular selection is disabled.

\subsection{Coverage vs.\ First-$K$ Initialization}
\label{app:submod-vs-firstk}

All main results use submodular initialization to construct the initial
MiA-Signature. To isolate the effect of this design choice, we compare it with a
simple First-$K$ initializer, which takes the session summaries associated with
the top-$K$ chunks from the step-$0$ query-only ranking. All other components are
kept fixed, including the retriever backbone, generator, and refinement budget.

\begin{table}[h]
\centering
\scriptsize
\setlength{\tabcolsep}{2.5pt}
\caption{\textbf{Coverage-aware vs.\ First-$K$ submodular initialization.}}
\label{tab:submod-vs-firstk}
\begin{tabular}{ll|ccc|cc|cc|c|cc}
\toprule
\multirow{2}{*}{Method} & \multirow{2}{*}{Init}
& \multicolumn{3}{c|}{DetectiveQA}
& \multicolumn{2}{c|}{NarrativeQA}
& \multicolumn{2}{c|}{NovelHopQA}
& NoCha
& \multicolumn{2}{c}{Avg.} \\
& & $R@10$ (EN/ZH) & Acc-EN & Acc-ZH
& $R@10$ & F1
& $R@10$ & F1
& PairAcc
& $R@10$ & Perf. \\
\midrule
\multirow{2}{*}{MiA-Emb}
  & First-$K$   & 44.7 / 42.7 & 69.3 & \underline{79.3} & 57.9 & 42.6 & \underline{37.0} & 37.7 & 58.7 & 46.2 & 53.3 \\
  & Submodular  & 44.7 / 42.7 & \underline{70.7} & 78.0 & \underline{59.5} & \underline{45.1} & 36.8 & \underline{38.5} & 58.7 & \underline{46.7} & \underline{54.2} \\
\midrule
\multirow{2}{*}{MiA-RAG (+MiA-Gen-14B)}
  & First-$K$   & 44.7 / 42.7 & 70.7 & 70.7 & 57.9 & 47.2 & \underline{37.0} & 27.7 & 47.6 & 46.2 & 48.3 \\
  & Submodular  & 44.7 / 42.7 & 70.7 & \underline{73.3} & \underline{59.5} & \underline{48.0} & 36.8 & \underline{27.9} & 47.6 & \underline{46.7} & \underline{48.9} \\
\midrule
\multirow{2}{*}{MiA-RAG (+DS-V3.2)}
  & First-$K$   & 44.7 / 42.7 & 74.0 & \underline{80.7} & 57.9 & 42.7 & \underline{37.0} & \underline{39.6} & 63.5 & 46.2 & 55.8 \\
  & Submodular  & 44.7 / 42.7 & \underline{74.7} & 80.0 & \underline{59.5} & \underline{42.8} & 36.8 & 38.7 & \underline{65.1} & \underline{46.7} & \underline{56.0} \\
\bottomrule
\end{tabular}
\end{table}

\paragraph{Findings.}
First-$K$ and Coverage-aware submodular use the same step-$0$ query-only candidate pool, so the initial evidence frontier is unchanged. Their difference lies in which high-level summaries are selected to form $\sigma_0$. Since the second retrieval pass is conditioned on both the query and this signature, different initializations lead to different query--signature retrieval distributions.

Coverage-aware submodular gives a small but consistent improvement in average $R@10$ across all three signature-based variants, and also improves average task performance in each case. The clearest gain appears on NarrativeQA, where the activated context is broad and redundant; selecting summaries for chunk coverage is therefore more useful than simply taking the first $K$ summaries from the ranking.

The effects are smaller and sometimes mixed on DetectiveQA, NovelHopQA, and NoCha. This is expected: when the activated region is narrower, or when the answer depends on precise local distinctions, the First-$K$ submodular variant can already provide a reasonable signature. Overall, the ablation shows that adding coverage-aware terms to the submodular objective is a modest but reliable improvement for static RAG, and justifies our default of using First-$K$ submodular initialization in the agent, where later refinement steps can compensate for a simpler initial objective.

\section{Retriever Mechanism}\label{app:retriever_mech}

\begin{figure}
    \centering
    \includegraphics[width=0.8\linewidth]{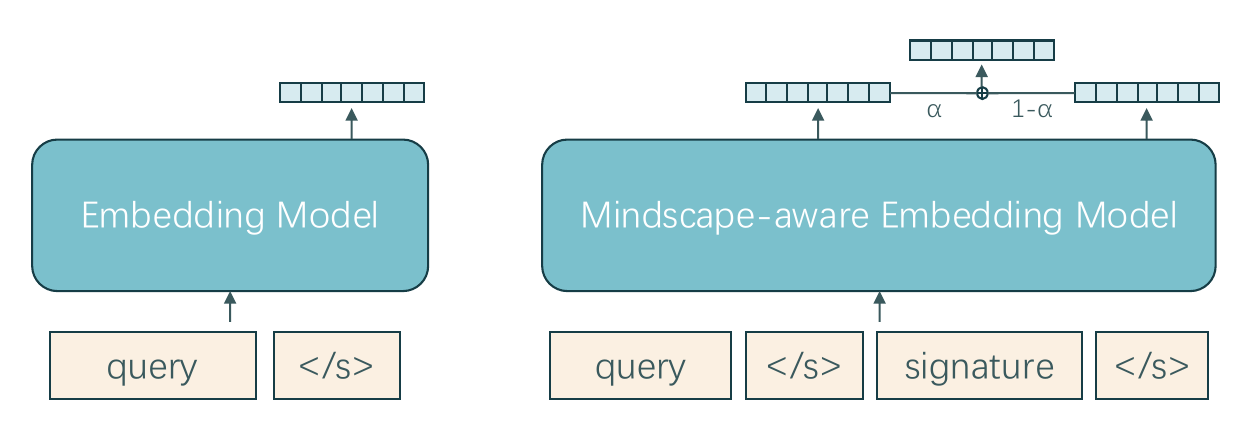}
    \caption{How the query-only embedding model and the mindscape-aware embedding model work when encoding the query end.}
    \label{fig:emb_work}
\end{figure}


We employ two types of retrievers to search chunks (evidences), utilizing either the query alone or the combination of the query and a signature. The first is referred to as a query-only retriever, which encodes the query using conventional last-token pooling; SFT-Embedding and Qwen3-Embedding belong to this category. The second is a mindscape-aware retriever (MiA-Embedding), which takes both the query and the signature as input following Eq.~\ref{eq:mia-score}. It captures two sources of information via interpolation: query-only information and signature-situated query information. This combination is effective because the causal attention mechanism in the decoder-structured retriever masks the signature tokens when processing the first </s> token. An illustration of the two retriever types is provided in Fig.~\ref{fig:emb_work}. Note that interpolation is applied only when encoding the query–signature pair; the chunk encoding process remains identical to that of the query-only retrievers.

\section{Dataset Construction and Statistics}
\label{app:datasets}

\subsection{Series Aggregation Details}
\label{app:datasets:series}

We aggregate books from original benchmarks to form coherent multi-volume series. For \textbf{DetectiveQA}, we group 13 novels into Miss Marple and Hercule Poirot series. For \textbf{NarrativeQA}, we select 37 books to form 11 series based on sequential arcs or shared protagonists. Table~\ref{tab:series-stats} summarizes the aggregation details.

\begin{table}[h]
\centering
\small
\caption{Statistics of aggregated series for DetectiveQA and NarrativeQA.}
\label{tab:series-stats}
\begin{tabular}{llcc}
\toprule
\textbf{Benchmark} & \textbf{Series / Arc} & \textbf{\#Books} & \textbf{\#Questions} \\
\midrule
DetectiveQA & Miss Marple (A. Christie) & 8 & 81 \\
            & Hercule Poirot (A. Christie) & 5 & 69 \\
\midrule
NarrativeQA & Anne of Green Gables & 4 & 42 \\
            & Balzac, La Com\'edie Humaine & 6 & 28 \\
            & Sherlock Holmes (A. C. Doyle) & 3 & 35 \\
            & Other (Indiana Jones, Star Wars, etc.) & 24 & 295 \\
\midrule
\textbf{Total} & \textbf{13 Series} & \textbf{50} & \textbf{550} \\
\bottomrule
\end{tabular}
\end{table}

\subsection{Single-Book vs. Series-Book Control Details}
\label{app:datasets:single-vs-series}

As a sanity check that series-book indexing is strictly harder than single-book indexing, we run a retrieval-side control with all other pipeline components held fixed. We use the same query-only retriever (SFT-Emb-8B) and compare two indexing granularities: \textbf{Single-Book}, where only the gold book is indexed and each question is answered against its own book; and \textbf{Series-Book}, where all books of a series are merged into a single document and retrieval is performed over the merged index. Table~\ref{tab:single-vs-series} reports retrieval recall under the two settings.

\subsection{On Full Global-Summary Baselines}
\label{app:datasets:global-summary-baseline}

Prior MiA-RAG~\cite{miarag} conditions retrieval and generation on a document-level global summary. This differs from the \textbf{MiA-RAG} system in this paper, where the conditioning signal is a query-induced MiA-Signature rather than a pre-existing document summary.

We do not include the prior full global-summary baseline in the series-book setting because the required summary is not well defined. A summary of the merged series would mix plots, characters, and events from multiple books, introducing semantic interference. A summary of only the gold book would require knowing the target book before retrieval, which leaks information.
Our setting instead tests whether the system can identify the query-relevant region from an overcomplete memory space. MiA-Signature is designed for this regime because it is induced by the query and does not assume access to a reliable document-level summary in advance.
\begin{table}[h]
\centering
\small
\caption{Query-only retrieval recall (\%) under single-book vs.\ series-book indexing, using SFT-Emb-8B with identical chunking and retrieval budget. Merging books of a series into a single index consistently lowers $R@k$, confirming that cross-book context acts as semantic interference rather than useful side information.}
\label{tab:single-vs-series}
\begin{tabular}{l|cc|cc}
\toprule
\multirow{2}{*}{Benchmark} & \multicolumn{2}{c|}{Single-Book} & \multicolumn{2}{c}{Series-Book} \\
 & $R@5$ & $R@10$ & $R@5$ & $R@10$ \\
\midrule
DetectiveQA (EN) & 30.0 & 40.0 & 29.3 & 38.0 \\
DetectiveQA (ZH) & 40.0 & 56.0 & 34.0 & 48.0 \\
NarrativeQA     & 48.2 & 61.0 & 44.0 & 55.5 \\
\bottomrule
\end{tabular}
\end{table}

The drop from Single-Book to Series-Book indicates that merging related books introduces additional semantic interference for query-only retrieval. This control motivates our use of the series-book setting in the main experiments: it better tests whether a memory interface can identify the query-relevant region before chunk-level matching. We therefore report all main-paper results under the harder series-book setting.

\begin{table}[h]
\centering
\scriptsize
\setlength{\tabcolsep}{3pt}
\caption{\textbf{Query-rewrite ablation.} All variants use an evolving
signature and differ only in whether the query is rewritten at each step;
the four blocks vary what the generator receives at answer time.}
\label{tab:coevo-ablation}

\resizebox{\textwidth}{!}{
\begin{tabular}{ll|cc|c|c|cc|c}
\toprule
\multirow{2}{*}{Variant}
& \multirow{2}{*}{Query rewrite}
& \multicolumn{2}{c|}{DetectiveQA Acc}
& NarrativeQA
& NovelHopQA
& \multicolumn{2}{c|}{NoCha}
& Avg. \\
& & EN & ZH & F1 & F1 & Acc & PairAcc & Perf. \\
\midrule
\multirow{2}{*}{chunks only}
& $\checkmark$ & 68.7 & 81.3 & \textbf{45.3} & 37.7 & 80.2 & 61.9 & 55.0 \\
& $\times$     & 70.0 & 78.0 & 44.0 & \textbf{38.7} & 80.2 & 60.3 & 54.3 \\
\midrule
\multirow{2}{*}{+sig}
& $\checkmark$ & \textbf{76.7} & 82.0 & 44.9 & 35.7 & 83.3 & 68.3 & \textbf{57.1} \\
& $\times$     & 69.3 & 80.0 & 43.8 & 37.1 & 84.1 & 68.3 & 56.0 \\
\midrule
\multirow{2}{*}{+evidence}
& $\checkmark$ & 73.3 & 80.7 & 43.3 & 35.9 & 83.3 & 66.7 & 55.7 \\
& $\times$     & 73.3 & \textbf{86.0} & 43.6 & 36.2 & 84.1 & 68.3 & 56.9 \\
\midrule
\multirow{2}{*}{+sig+evidence}
& $\checkmark$ & 73.3 & 80.0 & 44.3 & 35.0 & \textbf{85.7} & \textbf{71.4} & 56.8 \\
& $\times$     & 71.3 & 82.0 & 43.0 & 35.6 & 84.1 & 68.3 & 55.9 \\
\midrule
\multicolumn{2}{l|}{\textbf{Perf.\ $\Delta$ for +sig+evidence ($\checkmark-\times$)}}
& \textbf{+2.0} & \textbf{$-$2.0}
& \textbf{+1.3}
& \textbf{$-$0.6}
& \textbf{+1.6} & \textbf{+3.1}
& \textbf{+0.9} \\
\bottomrule
\end{tabular}
}
\end{table}

\section{Query-Rewrite Ablation}
\label{app:coevo-ablation}

While $\sigma_t$ always evolves with newly retrieved evidence, we ablate
whether the query $q_t$ should also be rewritten at each step.
Table~\ref{tab:coevo-ablation} reports final performance under the four
answer-time interfaces, toggling query rewriting on and off.

\paragraph{Findings.}
Rewriting helps when refinement should narrow the search---most clearly on NarrativeQA and NoCha, where partial evidence can be turned into a more specific follow-up query---but is not universally beneficial. NovelHopQA is the exception: keeping the query fixed yields higher F1, since multi-hop questions benefit from preserving parallel evidence paths rather than specializing around the first evidence found. Accordingly, our main experiments keep $q_t$ fixed on NovelHopQA and rewrite it elsewhere.
Query rewriting and signature evolution therefore play distinct roles. The signature carries the evolving global memory state; query rewriting only controls how narrowly the next retrieval step is posed, and we treat it as a benchmark-dependent control rather than a core mechanism.

\section{Case Study}
\label{app:case-study}

\definecolor{sigblue}{HTML}{E3F2FD}          
\definecolor{sigframe}{HTML}{1E6091}         
\definecolor{agentgreen}{HTML}{E8F5E9}       
\definecolor{agentframe}{HTML}{4F7C4A}       
\definecolor{neutralbg}{HTML}{F7F7F7}        
\definecolor{neutralframe}{HTML}{8A8A8A}     
\definecolor{goldbg}{HTML}{E8F5E9}
\definecolor{goldframe}{HTML}{2E7D32}
\definecolor{wrongbg}{HTML}{FDECEA}
\definecolor{wrongframe}{HTML}{B71C1C}
\definecolor{headergrey}{HTML}{F4F4F4}
\definecolor{submodbg}{HTML}{FFF8E1}         
\definecolor{submodframe}{HTML}{B89A2B}

We provide a DetectiveQA trace below. The compared systems retrieve locally plausible evidence for option~B, but fail to maintain the global identity
binding needed for the causal answer. MiA-Emb commits to the local surface reading, MiA-RAG's signature does not bind the hostess role to Charlotte-as-Letitia, and the agent without a signature has no state that preserves this binding across steps. In contrast, MiA-Agent updates the
signature once the binding is surfaced, allowing later retrieval and generation
to select the correct answer.
\begin{tcolorbox}[
  enhanced,
  breakable,
  colback=white,
  colframe=black!60,
  title={\textbf{Case study.} DetectiveQA--EN, Miss Marple merged series.},
  fonttitle=\footnotesize,
  coltitle=black,
  colbacktitle=headergrey,
  boxrule=0.6pt,
  arc=3pt,
  left=6pt,
  right=6pt,
  top=5pt,
  bottom=5pt,
  fontupper=\footnotesize,
]

\noindent\textbf{Question.}
\textit{The hostess's close friend, Dora, died on the second day of her birthday
party. What was the cause of her death?}

\smallskip
\noindent\textbf{Options.}
\begin{enumerate}[label=\Alph*., leftmargin=1.7em, itemsep=1pt, topsep=2pt]
  \item Rudi gives Dora poison to kill her in advance.
  \item Dora ingested poison.
  \item Someone entered the room and killed Dora while she was sleeping.
  \item The mistress substituted the sleeping pills that Dora took with poison.
  \textbf{(gold)}
\end{enumerate}

\medskip
\noindent\textbf{Why this case probes the signature.}
Option~B is supported by local chunks: the retriever surfaces that the pill Dora
took was not aspirin. Option~D requires a global binding from a different part
of the novel: the hostess ``Letitia'' Blacklock is in fact Charlotte
impersonating her dead sister, and Charlotte is the one who swapped the bedside
bottle. Since retrieval runs over the $8$-book Marple series, both the poisoning
subplot and the identity subplot can be activated. The central question is
whether the system can carry their binding across retrieval and generation.

\medskip
\noindent\textbf{Compared systems that answer incorrectly.}

{\centering
\begin{tcolorbox}[
  colback=wrongbg,
  colframe=wrongframe,
  boxrule=0.5pt,
  arc=2pt,
  left=6pt,
  right=6pt,
  top=4pt,
  bottom=4pt,
  width=0.96\linewidth,
  breakable
]
\textsc{MiA-Emb}. Predicts
\textcolor{wrongframe}{\textbf{B}}. No signature is provided to the generator;
the retrieved context contains the party scene and the non-aspirin pill, so the
model commits to the local surface reading.

\smallskip
\textsc{MiA-RAG}. Predicts
\textcolor{wrongframe}{\textbf{B}}. The generator receives a signature,
but the signature does not yet bind the hostess role to Charlotte-as-Letitia.

\smallskip
\textsc{ (Agent without signature) }. Predicts
\textcolor{wrongframe}{\textbf{B}}. The query is rewritten across steps, but no
global signature is carried to preserve the Charlotte--Letitia binding.
\end{tcolorbox}
\par}

\medskip
\noindent\textbf{\method{} (ours).}
Two refinement steps, confidence \textsc{medium} $\to$ \textsc{high}.

\begin{tcolorbox}[
  colback=submodbg,
  colframe=submodframe,
  boxrule=0.5pt,
  arc=2pt,
  title={\footnotesize\textbf{Step 0.} Initial signature construction},
  fonttitle=\footnotesize,
  coltitle=black,
  colbacktitle=submodbg!60,
  left=6pt,
  right=6pt,
  top=4pt,
  bottom=4pt,
  breakable
]
The query-only retriever $\mathcal{E}_1$ returns the top-$50$ chunks from the
merged Marple series. Their session summaries deduplicate to a candidate pool
with $|\mathcal{H}_0(q)|=10$. First-$K$ submodular selection keeps the top
$K=5$ summaries by query relevance.

\smallskip
\noindent\textbf{Selected summaries include:}
\begin{itemize}[leftmargin=1.2em,itemsep=1pt,topsep=2pt]
\item \emph{Miss Marple reveals that Charlotte Blacklock assumed her sister
Letitia's identity. Charlotte invited Dora Bunner, but Dora's forgetfulness
threatened Charlotte's secret.}
{\footnotesize $(q=0.89)$}

\item \emph{Poisoned aspirins that killed Dora Bunner were placed during the
birthday party; the positions of the participants are reconstructed.}
{\footnotesize $(q=0.84)$}

\item \emph{Dora Bunner was found dead after taking pills from Blacklock's
bedside; one remaining pill was not aspirin.}
{\footnotesize $(q=1.00)$}

\item Two additional top-ranked summaries, including one semantically related
Marple scene that acts as distractor context.
\end{itemize}

The resulting $\sigma_0$ activates both the identity subplot and the poisoning
subplot, but it has not yet committed the binding between the hostess identity
and the pill substitution.
\end{tcolorbox}

\begin{tcolorbox}[
  colback=agentgreen,
  colframe=agentframe,
  boxrule=0.5pt,
  arc=2pt,
  title={\footnotesize\textbf{Step 1.} \textsc{refine}, confidence \textsc{medium}},
  fonttitle=\footnotesize,
  coltitle=black,
  colbacktitle=agentgreen!45,
  left=6pt,
  right=6pt,
  top=4pt,
  bottom=4pt,
  breakable
]
\textbf{Query $q_0$.}
\textit{What was the cause of Dora's death?}

\smallskip
\begin{tcolorbox}[
  colback=neutralbg,
  colframe=neutralframe,
  boxrule=0.4pt,
  arc=1.5pt,
  left=5pt,
  right=5pt,
  top=3pt,
  bottom=3pt,
  breakable
]
\textbf{Evidence memory $E_1$ grounded in retrieval.}
\begin{itemize}[leftmargin=1.2em,itemsep=1pt,topsep=1pt]
\item Dora took aspirin from a bottle on Miss \emph{Letitia} Blacklock's
bedside; the remaining pill was not aspirin.
\item The poisoned aspirins were placed during Dora's birthday party; anyone
present could have switched them.
\item Charlotte Blacklock, posing as Letitia, killed Dora because Dora's
forgetfulness threatened Charlotte's secret identity.
\end{itemize}
\end{tcolorbox}

\smallskip
\textbf{Reason.}
$E_1$ establishes the mechanism, namely a non-aspirin pill, and places the
swap at the birthday party. However, the agent of the swap is not yet bound to
the hostess identity referred to by option~D. The agent therefore refines rather
than answers.
\end{tcolorbox}

{\centering
\begin{tcolorbox}[
  colback=neutralbg,
  colframe=neutralframe,
  boxrule=0.5pt,
  arc=2pt,
  left=6pt,
  right=6pt,
  top=4pt,
  bottom=4pt,
  width=0.96\linewidth,
  breakable
]
\textbf{Co-evolution at the end of Step~1.}
The agent updates $(q_0,\sigma_0,E_0)$ into $(q_1,\sigma_1,E_1)$.

\smallskip
\textbf{Rewritten query $q_1$.}
\textit{Who substituted the aspirin pills that Dora Bunner took, and what
specific poison caused her death?}

\smallskip
\begin{tcolorbox}[
  colback=sigblue,
  colframe=sigframe,
  boxrule=0.4pt,
  arc=1.5pt,
  left=5pt,
  right=5pt,
  top=3pt,
  bottom=3pt,
  breakable
]
\textbf{Refined signature $\sigma_1$ written by $M_{\mathrm{upd}}$.}
\textit{Miss Marple reveals the full truth at the parsonage. Charlotte
Blacklock, living under her dead sister Letitia's identity, saw her old friend
Dora Bunner as a threat due to Dora's increasing forgetfulness and talkativeness.
To protect her secret, Charlotte substituted the aspirin in her own bedside
bottle with poisoned pills, knowing Dora would take them.}
\end{tcolorbox}

The binding
``hostess $=$ Charlotte-as-Letitia $=$ agent of the swap''
is now encoded inside $\sigma_1$.
\end{tcolorbox}
\par}

\begin{tcolorbox}[
  colback=agentgreen,
  colframe=agentframe,
  boxrule=0.5pt,
  arc=2pt,
  title={\footnotesize\textbf{Step 2.} \textsc{answer}, confidence \textsc{high}},
  fonttitle=\footnotesize,
  coltitle=black,
  colbacktitle=agentgreen!45,
  left=6pt,
  right=6pt,
  top=4pt,
  bottom=4pt,
  breakable
]
Signature-guided retrieval $\mathcal{E}_2$ now retrieves under
$(q_1,\sigma_1)$. Miss Marple's parsonage reveal is promoted because $\sigma_1$
biases retrieval toward the identity-bound region, not merely because the
chunks mention Dora.

\smallskip
\begin{tcolorbox}[
  colback=neutralbg,
  colframe=neutralframe,
  boxrule=0.4pt,
  arc=1.5pt,
  left=5pt,
  right=5pt,
  top=3pt,
  bottom=3pt,
  breakable
]
\textbf{Evidence memory $E_2$ updated this step.}
\begin{itemize}[leftmargin=1.2em,itemsep=1pt,topsep=1pt]
\item Charlotte substituted the aspirin in her, that is, ``Letitia's'', bedside bottle with poisoned pills.
\item The specific poison was aconite.
\item Charlotte framed the killing as an act of mercy for Dora.
\end{itemize}
\end{tcolorbox}

\smallskip
\textbf{Reason.}
With $\sigma_1$ identifying the hostess as the agent of the swap, ``the mistress'' in option~D becomes bindable. Option~B remains locally true, but it is only the surface mechanism. Option~D gives the causal explanation. The agent answers.

\smallskip
\textbf{Final answer.}
\colorbox{goldbg}{\textcolor{goldframe}{\textbf{D}}}
\;$\checkmark$
\end{tcolorbox}

\end{tcolorbox}

\section{Prompt Templates}
\label{app:prompts}

\definecolor{promptbg}{HTML}{F5F5F5}
\definecolor{promptframe}{HTML}{616161}

\tcbset{
  promptbox/.style={
    enhanced,
    breakable,
    colback=promptbg,
    colframe=promptframe,
    boxrule=0.4pt,
    arc=2pt,
    left=5pt,
    right=5pt,
    top=4pt,
    bottom=4pt,
    fonttitle=\small,
    coltitle=black,
    colbacktitle=promptbg,
    listing only,
    listing options={
      basicstyle=\scriptsize\ttfamily,
      breaklines=true,
      breakatwhitespace=false,
      columns=fullflexible,
      keepspaces=true,
      showstringspaces=false,
      tabsize=2
    }
  }
}

We list the prompts used by \method{}. The \textsc{Session-Summary} prompt is
called once per sessionization window during offline preprocessing to
construct the high-level memory set $\mathcal{H}(D)$. The \textsc{Update}
prompt is called at each refinement step by $M_{\mathrm{upd}}$, and the
answer prompt is sent to $M_{\mathrm{gen}}$ once the agent decides to
answer. Placeholders are shown in \texttt{\{braces\}}.

\begin{tcblisting}{
  promptbox,
  title={\textsc{Session-Summary} Prompt}
}
You are provided with a raw text segment from a book (Part {idx}/{total}).
This segment consists of approximately 20 consecutive chunks combined.

<Raw_Text>
{raw_text}
</Raw_Text>

Please generate a **Detailed Narrative Summary** following these strict guidelines:

1. **Narrative Reconstruction**: Do not list events. Rewrite the content as a coherent story in the third person, past tense. It should read like a condensed version of the original text.
2. **Detail Preservation**:
   - Preserve specific **Character Names** and their relationships.
   - Keep key **Dialogues** that drive the plot.
   - Note specific **Locations** or setting changes.
3. **Noise Filtering**:
   - IGNORE any copyright notices, Project Gutenberg headers, page numbers, or tables of contents.
   - If the text starts or ends in the middle of a sentence, ignore the broken fragments and focus on the complete thoughts.
4. **Style**:
   - NO meta-commentary (e.g., do NOT say "The text describes...", "In this chunk...").
   - Directly tell the story.
5. **Length**: 50--100 words.

Output the summary directly.
\end{tcblisting}

\begin{tcblisting}{
  promptbox,
  title={\textsc{Update} Prompt}
}
[System]
You are a retrieval planning agent. You decide whether retrieved passages contain enough evidence to answer a question, and if not, co-refine the retrieval signature and search query.

You work with a mindscape-aware retriever guided by TWO evolving signals:
1. Signature: A short narrative summary --- the "compass" steering retrieval toward the right storyline, characters, and events.
2. Search Query: A rewritten question targeting specific missing evidence. The retriever combines both signals to find passages.

Information hierarchy:
- Session Summaries: high-level narrative overviews for orientation.
- Retrieved Passages: specific text chunks from the current retrieval step.
- Evidence Memory: running notes accumulated across all steps.

Decision process:
- Most evidence found -> ANSWER
- Key evidence missing + steps remaining -> REFINE
- LOW confidence + steps remaining -> REFINE
- Last step -> ANSWER

When action=REFINE:
- Output both a refined signature and a rewritten query.
- The signature targets the missing evidence and relevant storyline.
- The query targets concrete missing evidence.

Output format:
<evidence_memory> bullet list of key findings </evidence_memory>
<confidence>HIGH/MEDIUM/LOW</confidence>
<thought>reasoning: what is found, missing, chosen action</thought>
<action>ANSWER or REFINE</action>
<refined_signature>updated narrative compass, only if REFINE</refined_signature>
<rewritten_query>targeted search query, only if REFINE</rewritten_query>

[User]
Question:
{question}

Options:
{options_str}

Step {step}/{max_steps} | {remaining_steps_hint}
Current signature:
{signature}

Current search query:
{current_query}

Session summaries:
{summaries_text}

Evidence memory:
{evidence_memory}

Retrieved passages:
{chunks_text}

{history_section}

Do the passages above, understood in the context of the session summaries and
your current signature, contain enough evidence to answer the question?
\end{tcblisting}

\paragraph{Answer-time input variants.}
All answer-time variants include the final retrieved chunks. The following
table only shows which additional memory states are prepended to the generator
context.

\begin{center}
\small
\setlength{\tabcolsep}{5pt}
\begin{tabular}{ll}
\toprule
Variant & Generator context \\
\midrule
Chunks & \texttt{\{context\}} \\
Chunks + \textsc{Sig.} &
\texttt{\{signature\}} $+$ \texttt{\{context\}} \\
Chunks + \textsc{Evi.} &
\texttt{\{evidence\_memory\}} $+$ \texttt{\{context\}} \\
Chunks + \textsc{Sig.} + \textsc{Evi.} &
\texttt{\{signature\}} $+$ \texttt{\{evidence\_memory\}} $+$ \texttt{\{context\}} \\
\bottomrule
\end{tabular}
\end{center}

\begin{tcblisting}{
  promptbox,
  title={Answer Prompts and Dataset-Specific Output Formats}
}
--- DetectiveQA ---

[System]
You are a helpful assistant.

[User]
{answer_context}

Please answer the question based on the current novel content:
{question}
{options_str}

Please strictly follow the format {"answer":"x","reasoning":"xxx"}. The answer field should only contain A, B, C, or D.


------------------------------------------------------------

--- Open QA (NarrativeQA, NovelHopQA) ---

[System] You are an expert reading comprehension assistant. Analyze provided passages to answer the question as concisely as possible, using a single phrase if possible. Do not provide any explanation.

[User] ... {context} ...
Now, answer the question based on the story as concisely as you can, using a single phrase if possible. Do not provide any explanation.
Question: {question}
Answer:


------------------------------------------------------------

--- Claim Verification (NoCha) ---

[System] You are an expert reading comprehension assistant. You verify whether statements about novels are TRUE or FALSE based on provided evidence.

[User] ... {context} ...
<statement>{claim}</statement>
<question>Based on the context provided, is the above statement TRUE or FALSE?</question>

Answer TRUE if the statement is true in its entirety based on the context provided.
Answer FALSE if any part of the statement is false based on the context provided.

First provide an explanation in at most one paragraph, then your final answer:
<explanation>YOUR EXPLANATION</explanation>
<answer>YOUR ANSWER</answer>
\end{tcblisting}

\section{Limitations}
\label{sec:limitations}

Our results show that MiA-Signatures offer an effective memory interface for long-context narrative understanding, especially when evidence is dispersed across a large source. Still, our experiments are centered on literary and narrative domains where memory naturally forms chapter- or session-level units; whether the same activation--signature formulation transfers to code repositories, scientific literature, or multimodal interaction remains to be tested. The current signature construction is also training-free and based on submodular selection over precomputed summaries, which keeps the method modular but does not optimize the signature end-to-end with the retriever, generator, or task objective. Finally, MiA-Signature should be understood as a global-structure prior rather than a replacement for local evidence: it helps when answers require synthesis across dispersed context, but can be unnecessary or distracting when the answer is already locally supported. Adaptive control
over when to expose the signature to the generator remains future work.
\clearpage

\newpage
\end{document}